%% file: paper_for_review.tex
\documentclass[11pt,a4paper]{article}
\usepackage[T1]{fontenc}
\usepackage[utf8]{inputenc}
\usepackage[hyperref]{acl2018}
\usepackage{times}
\usepackage{epsfig}
\usepackage{booktabs}
\usepackage{graphicx}
\usepackage{algorithm}
\usepackage{tabulary}
\usepackage[noend]{algpseudocode}
\usepackage{subcaption}
\usepackage{paralist}
\makeatletter
\def\BState{\State\hskip-\ALG@thistlm}
\makeatother
\usepackage[page]{appendix}
\usepackage{comment}
\usepackage{amsmath}
\usepackage{amssymb}
\usepackage{mathtools}
\usepackage{enumitem}
\setlist{nosep}
\usepackage{bm} 

\DeclareMathOperator\arctanh{arctanh}

\usepackage{latexsym}

\usepackage{url}
\aclfinalcopy 
\def\aclpaperid{1245} 

\title{Attacking Visual Language Grounding with Adversarial Examples:\\ A Case Study on Neural Image Captioning}

\author{Hongge Chen\textsuperscript{1}\textsuperscript{*}, \enskip Huan Zhang\textsuperscript{23}\textsuperscript{*}, \enskip Pin-Yu Chen\textsuperscript{3}, \enskip Jinfeng Yi\textsuperscript{4}, \enskip {\normalfont and} \enskip Cho-Jui Hsieh\textsuperscript{2} \\
  \textsuperscript{1}MIT, Cambridge, MA 02139, USA\\
  \textsuperscript{2}UC Davis, Davis, CA 95616, USA \\
  \textsuperscript{3}IBM Research, NY 10598, USA \\
  \textsuperscript{4}JD AI Research, Beijing, China\\
  {\tt\small chenhg@mit.edu, ecezhang@ucdavis.edu} \\
  {\tt\small pin-yu.chen@ibm.com, yijinfeng@jd.com, chohsieh@ucdavis.edu}\\
  \textsuperscript{*}{\normalsize Hongge Chen and Huan Zhang contribute equally to this work}\\
  }

\begin{document}
\maketitle

\begin{abstract}
Visual language grounding is widely studied in modern neural image captioning systems, which typically adopts an encoder-decoder framework consisting of two principal components: a convolutional neural network (CNN) for image feature extraction and a recurrent neural network (RNN) for language caption generation. 
To study the robustness of language grounding to adversarial perturbations in machine vision and perception, 
we propose \textbf{Show-and-Fool}, a novel algorithm for crafting adversarial examples in neural image captioning. The proposed algorithm  provides  
two evaluation approaches, which check whether neural image captioning systems can be mislead to output some randomly chosen captions or keywords. 
Our extensive experiments show that our algorithm can successfully craft visually-similar adversarial examples with randomly targeted captions or keywords, and the adversarial examples can be made highly transferable to other image captioning systems. Consequently, our approach leads to new robustness implications of neural image captioning and novel insights in visual language grounding.

\end{abstract}


\section{Introduction}
In recent years, language understanding grounded in machine vision and perception has made remarkable progress in natural language processing (NLP) and artificial intelligence (AI), such as image captioning and visual question answering. 
Image captioning is a multimodal learning task and has been used to study the interaction between language and vision models \cite{shekhar2017foil}. It takes an image as an input and generates a language caption that best describes its visual contents, and has many important applications such as developing image search engines with complex natural language queries, building AI agents that can see and talk, and promoting equal web access for people who are blind or visually impaired. Modern image captioning systems typically adopt an encoder-decoder framework composed of two principal modules: a convolutional neural network (CNN) as an encoder for image feature extraction and a recurrent neural network (RNN) as a decoder for caption generation. 
This CNN+RNN architecture includes popular image captioning models  such as Show-and-Tell ~\cite{DBLP:conf/cvpr/VinyalsTBE15}, Show-Attend-and-Tell~\cite{DBLP:conf/icml/XuBKCCSZB15} and NeuralTalk~\cite{DBLP:conf/cvpr/KarpathyL15}.
\begin{figure}[t]
        \centering
        \begin{subfigure}[b]{0.93\linewidth}
            \includegraphics[width=\textwidth]{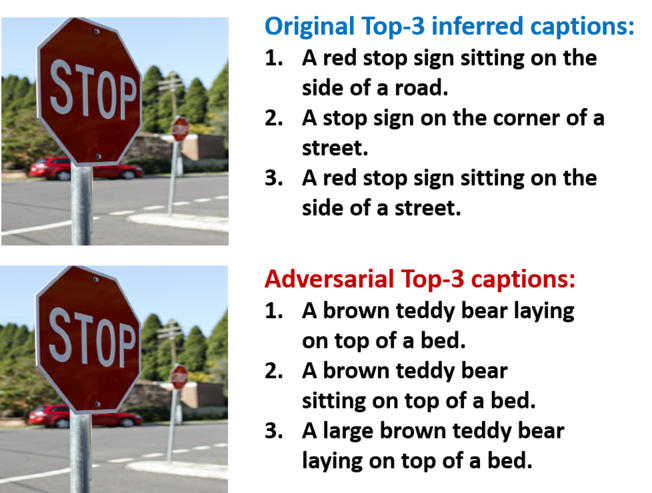}
        \end{subfigure}
         \centering
         \begin{subfigure}[b]{0.93\linewidth}
             \includegraphics[width=\textwidth]{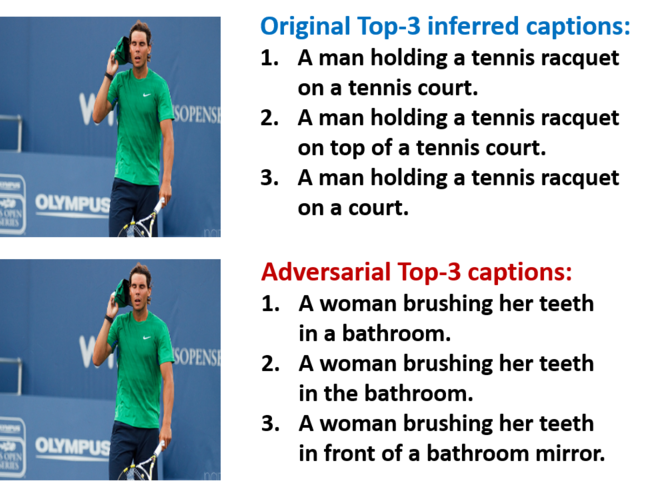}
         \end{subfigure}
        \caption{Adversarial examples crafted by Show-and-Fool using the targeted caption method. The target captioning model is Show-and-Tell~\cite{DBLP:conf/cvpr/VinyalsTBE15},
the original images are selected from the MSCOCO validation set, and the
targeted captions are randomly selected from the top-1 inferred caption of other  validation images.}   
       \label{Fig_show_fool}
    \end{figure}

Recent studies have highlighted the vulnerability of CNN-based image classifiers to adversarial examples:  adversarial perturbations to benign images can be easily crafted to mislead a well-trained classifier, leading to visually indistinguishable adversarial examples to human \cite{szegedy2013intriguing,goodfellow2014explaining}. 
In this study, we investigate a more challenging problem in visual language grounding domain that evaluates the robustness of multimodal RNN in the form of a CNN+RNN architecture, 
and use neural image captioning as a case study. 
Note that crafting adversarial examples in image captioning tasks is strictly harder than in well-studied image classification tasks, due to the following reasons: (i) class attack v.s. caption attack: unlike classification tasks where the class labels are well defined, the output of image captioning is a set of top-ranked captions. Simply treating different captions as distinct classes will result in an enormous number of classes that can even precede the number of training images. In addition, semantically similar captions can be expressed in different ways and hence should not be viewed as different classes; and (ii)~CNN v.s. CNN+RNN: attacking RNN models is significantly less well-studied than attacking CNN models. The CNN+RNN architecture is unique and beyond the scope of adversarial examples in CNN-based image classifiers.

In this paper, we tackle the aforementioned challenges by proposing a novel algorithm called \textit{Show-and-Fool}. 
We formulate the process of crafting adversarial examples in neural image captioning systems as optimization problems with novel objective functions designed to adopt the CNN+RNN architecture. Specifically, our objective function is a linear combination of the distortion between benign and adversarial examples as well as some carefully designed loss functions. 
The proposed Show-and-Fool algorithm provides two approaches to craft adversarial examples in neural image captioning under different scenarios: 
\begin{compactenum}
\item \textbf{Targeted caption method:} Given a targeted caption, craft adversarial perturbations to any image such that its generated caption matches the targeted caption. 
\item \textbf{Targeted keyword method:} Given a set of keywords, craft adversarial perturbations to any image such that its generated caption  contains the specified keywords. The captioning model has the freedom to make sentences with target keywords \textit{in any order}.
\end{compactenum}
As an illustration, Figure \ref{Fig_show_fool} shows an adversarial example crafted by Show-and-Fool using the targeted caption method. The adversarial perturbations are visually imperceptible while can successfully mislead Show-and-Tell to generate the targeted captions. Interestingly and perhaps surprisingly, our results pinpoint the Achilles heel of the language and vision models used in the tested image captioning systems. Moreover, the adversarial examples in neural image captioning highlight the inconsistency in visual language grounding between humans and machines, suggesting
a possible weakness of current machine vision and perception machinery.
Below we highlight our major contributions:


\begin{itemize}[leftmargin=*]
\item We propose \textit{Show-and-Fool}, a novel optimization based approach to crafting adversarial examples in image captioning. We provide two types of adversarial examples, targeted caption and targeted keyword, to analyze the robustness of neural image captioners. To the best of our knowledge, this is the very first work on crafting adversarial examples for image captioning.

\item We propose powerful and generic loss functions that can craft adversarial examples and evaluate the robustness of the encoder-decoder pipelines in the form of a CNN+RNN architecture. 
In particular, our loss designed for targeted keyword attack only requires the adversarial caption to contain a few specified keywords; and we allow the neural network to \textit{make meaningful sentences with these keywords on its own}.

\item 
We conduct extensive experiments on the MSCOCO dataset. Experimental results show that our targeted caption method attains a 95.8\% attack success rate when crafting adversarial examples with randomly assigned captions. In addition, our targeted keyword attack yields an even higher success rate. We also show that attacking CNN+RNN models is inherently different and more challenging than  only attacking CNN models.

\item We also show that Show-and-Fool can produce highly transferable adversarial examples: an adversarial image generated for fooling Show-and-Tell can also fool other image captioning models, leading to new robustness implications of neural image captioning systems.

\end{itemize}


\section{Related Work}

In this section, we review the existing work on visual language grounding, with a focus on neural image captioning. We also review related work on adversarial attacks on CNN-based image classifiers. Due to space limitations, we defer the second part to the supplementary material.

Visual language grounding represents a family of multimodal tasks that bridge visual and natural language understanding. Typical examples include image and video captioning~\cite{DBLP:conf/cvpr/KarpathyL15,DBLP:conf/cvpr/VinyalsTBE15,donahue2015long,DBLP:conf/acl/PasunuruB17,DBLP:conf/naacl/VenugopalanXDRM15}, visual dialog~\cite{das2017visual,de2017guesswhat}, visual question answering~\cite{antol2015vqa,fukui2016multimodal,lu2016hierarchical,zhu2017uncovering}, visual storytelling~\citep{huang2016visual}, natural question generation~\citep{mostafazadeh2017image,mostafazadeh2016generating}, and image generation from captions~\cite{mansimov2015generating,reed2016generative}. In this paper, we focus on studying the robustness of neural image captioning models, and believe that the proposed method also sheds lights on robustness evaluation for other visual language grounding tasks using a similar multimodal RNN architecture. 

Many image captioning methods based on deep neural networks (DNNs) adopt a multimodal RNN framework that first uses a CNN model as the encoder to extract a visual feature vector, followed by a RNN model as the decoder for caption generation. Representative works under this framework include \cite{DBLP:conf/cvpr/ChenZ15,devlin2015language,DBLP:conf/cvpr/DonahueHGRVDS15,DBLP:conf/cvpr/KarpathyL15,DBLP:journals/corr/MaoXYWY14a,DBLP:conf/cvpr/VinyalsTBE15,
DBLP:conf/icml/XuBKCCSZB15,DBLP:conf/nips/YangYWCS16,liu2017attention,liu2017semantic}, which are mainly differed by the underlying CNN and RNN architectures, and whether or not the attention mechanisms are considered. 
Other lines of research generate image captions using semantic information or via a compositional approach~\cite{fang2015captions,gan2017semantic,tran2016rich,jia2015guiding,wu2016value,you2016image}.  

The recent work in \cite{shekhar2017foil} touched upon the robustness of neural image captioning for language grounding by showing its insensitivity to one-word (foil word) changes in the language caption, which corresponds to the \textit{untargeted attack} category in adversarial examples. In this paper, we focus on the more challenging \textit{targeted attack} setting that requires to fool the captioning models and enforce them to generate pre-specified captions or keywords.

\section{Methodology of Show-and-Fool}
\subsection{Overview of the Objective Functions}
We now formally introduce our approaches to crafting adversarial
examples for neural image captioning. 
The problem of finding an adversarial example for a given image $I$ can be cast as the following optimization problem:
\begin{align}
\min_\delta\  &   \ c\cdot \text{loss}(I+\delta)+\|\delta\|_2^2\nonumber\\
\text{s.t. }\  &  \ I+\delta\in [-1,1]^n. \label{opt}
\end{align}
Here $\delta$ denotes the adversarial perturbation to $I$. $\|\delta\|_2^2=\|(I+\delta)-I\|_2^2$ is an $\ell_2$ distance metric between the original image and the adversarial image.  $\text{loss}(\cdot)$ is an attack loss function which takes different forms in different attacking settings. We will provide the explicit expressions in Sections \ref{sec:exact}  and \ref{sec:keyword}.
The term $c>0$ is a pre-specified regularization constant. Intuitively, with larger $c$, the attack is more likely to succeed but at the price of higher distortion on $\delta$. In our algorithm, we use a binary search strategy to select $c$. The box constraint on the image $I\in[-1,1]^n$ ensures that the adversarial example $I+\delta\in[-1,1]^n$ lies within a valid image space.

For the purpose of efficient optimization, we convert the constrained minimization problem in (\ref{opt}) into an unconstrained minimization problem by
introducing two new variables $y \in\mathbb{R}^n $ and $w \in\mathbb{R}^n$ such that
\begin{equation*}\label{eq:transform}
y=\arctanh(I) \ \text{ and } \ w=\arctanh(I+\delta)-y,
\end{equation*}where $\arctanh$ denotes the inverse hyperbolic tangent function and is applied element-wisely. Since $\tanh(y_i+w_i)\in[-1,1]$, the transformation will  automatically satisfy  the box constraint. Consequently, the constrained optimization problem in (\ref{opt}) is equivalent to
\begin{eqnarray}
\label{eq:general_loss}
&\min_{w\in\mathbb{R}^n}&c\cdot \text{loss}(\tanh(w+y))\\
&&+\|\tanh(w+y)-\tanh(y)\|_2^2 \nonumber.
\end{eqnarray}
In the following sections, we present our designed loss functions for different attack settings.

\subsection{Targeted Caption Method}\label{sec:exact}

Note that a targeted caption is denoted by
$$S=(S_1,\ S_2,\ ...,\ S_t,\ ...,\ S_N),$$ where $S_t$ indicates the index of the $t$-th word in the vocabulary list $\mathcal{V}$, $S_1$ is a start symbol and $S_N$ indicates the end symbol. 
$N$ is the length of caption $S$, which is not fixed but does not exceed a predefined maximum caption length. To encourage the neural image captioning system to output the targeted caption $S$, one needs to ensure the log probability of the caption $S$ conditioned on the image $I+\delta$ attains the maximum value among all possible captions, that is,
\begin{equation}\label{minimization}
\log P(S|I+\delta) = \max_{S' \in \Omega} \log P(S'|I+\delta),
\end{equation}
where $\Omega$ is the set of all possible captions.
It is also common to apply the chain rule to the joint probability and we have
\begin{equation*}
\log P(S'|I+\delta)=\sum_{t=2}^{N}\log P(S'_t|I+\delta,S'_1,...,S'_{t-1}).
\end{equation*}

In neural image captioning networks, $p(S'_t|I+\delta,S'_1,...,S'_{t-1})$ is usually computed by a RNN/LSTM cell $f$, with its hidden state $h_{t-1}$ and input $S^\prime_{t-1}$:
\begin{equation}
\label{eq:rnn}
z_t = f(h_{t-1}, S^\prime_{t-1}) \text{ and } p_t = \text{softmax}(z_t),
\end{equation}
where $z_t\coloneqq[z_t^{(1)}, z_t^{(2)},..., z_t^{(|\mathcal{V}|)}]\in \mathbb{R}^{|\mathcal{V}|}$ is a vector of the \textit{logits} (unnormalized probabilities) for each possible word in the vocabulary. The vector $p_t$ represents a probability distribution on $\mathcal{V}$ with each coordinate $p_t^{(i)}$ defined as:
$$p_t^{(i)} \coloneqq P(S'_{t}=i|I+\delta,S'_1,...,S'_{t-1}).$$ 
Following the definition of softmax function:
\begin{equation}
P(S'_t|I+\delta,S'_1,...,S'_{t-1})=\exp(z_t^{(S'_t)})/\sum_{i \in \mathcal{V}}\exp(z_t^{(i)}). \nonumber
\end{equation}

Intuitively, to maximize the targeted caption's probability, we can directly use its negative log probability (\ref{eq:softmaxloss}) as a loss function. The inputs of the RNN are the first $N-1$ words of the targeted caption $(S_1,S_2,...,S_{N-1})$. 
\begin{equation}\label{eq:softmaxloss}
\begin{gathered}
\text{loss}_{S, \text{log-prob}}(I+\delta)=-\log P(S|I+\delta)\\=-\sum_{t=2}^{N}\log P(S_t|I+\delta,S_1,...,S_{t-1}). 
\end{gathered}
\end{equation}
Applying (\ref{eq:softmaxloss}) to (\ref{eq:general_loss}), the formulation of targeted caption method given a targeted caption $S$ is: 
\begin{align*}
\min_{w\in \mathbb{R}^n} &c \cdot \text{loss}_{S, \text{log prob}}(\tanh(w+y)) \nonumber \\ 
&+\ \|\tanh(w+y)-\tanh(y)\|_2^2.
\end{align*}

Alternatively, using the definition of the softmax function,
\begin{align}
\log P(S'|I+\delta)&=\sum_{t=2}^N[z_t^{(S'_t)}-\log(\sum_{i \in \mathcal{V}} \exp(z_t^{(i)}))] \nonumber \\
&=\sum_{t=2}^Nz_t^{(S'_t)}-\text{constant,}
\end{align}
(\ref{minimization}) can be simplified as 
\begin{equation*}
\log P(S|I+\delta) \propto \sum_{t=2}^N z_t^{(S_t)}=\max_{S' \in \Omega}\sum_{t=2}^N z_t^{(S'_t)}. 
\end{equation*}

Instead of making each $z_t^{(S_t)}$ as large as possible,
it is sufficient to require the target word $S_t$ to attain the largest (top-1) logit (or probability) among all the words in the vocabulary at position $t$. In other words, we aim to minimize the difference between the maximum logit except $S_t$, denoted by $\max_{k \in \mathcal{V}, k\neq S_t}\{z_t^{(k)}\}$, and the logit of $S_t$, denoted by $z_t^{(S_t)}$. We also propose a ramp function on top of this difference as the final loss function:
\begin{equation}\label{eq:logits loss}
\!\!\!\text{loss}_{S, \text{logits}}(I+\delta)=\!\!\sum_{t=2}^{N-1}\max\{-\epsilon, \max_{k\neq S_t}\{z^{(k)}_t\}-z_t^{(S_t)}\},
\end{equation}
where $\epsilon>0$ is a confidence level accounting for the gap between $\max_{k\neq S_t}\{z^{(k)}_t\}$ and $z_t^{(S_t)}$. 
When $z_t^{(S_t)} > \max_{k\neq S_t}\{z^{(k)}_t\} + \epsilon$, the corresponding term in the summation will be kept at $-\epsilon$ and does not contribute to the gradient of the loss function, encouraging the optimizer to focus on minimizing other terms where  $z_t^{(S_t)}$ is not large enough.

Applying the loss (\ref{eq:logits loss}) to (\ref{opt}), the final formulation of targeted caption method given a targeted caption $S$ is 
\begin{equation*}
\begin{gathered}
\min_{w\in \mathbb{R}^n} c\cdot\sum_{t=2}^{N-1}\max\{-\epsilon, \max_{k\neq S_t}\{z^{(k)}_t\}-z_t^{(S_t)}\}\\+\ \|\tanh(w+y)-\tanh(y)\|_2^2.
\end{gathered}
\end{equation*}

We note that \cite{carlini2017towards} has reported that in CNN-based image classification, using logits in the attack loss function can produce better adversarial examples than using probabilities, especially when the target network deploys some gradient masking schemes such as defensive distillation \cite{papernot2016distillation}. Therefore,  we provide both logit-based and probability-based attack loss functions for neural image captioning.

\subsection{Targeted Keyword Method}\label{sec:keyword}

In addition to generating an exact targeted caption by perturbing the input image,  we offer an intermediate option that aims at generating captions with specific keywords, denoted by $\mathcal{K}\coloneqq\{K_1, \cdots, K_M\} \subset \mathcal{V}$. Intuitively, finding an adversarial image generating a caption with specific keywords might be easier than generating an exact caption, as we allow more degree of freedom in caption generation. However, as we need to ensure a valid and meaningful inferred caption, finding an adversarial example with specific keywords in its caption is difficult in an optimization perspective. Our target keyword method can be used to investigate the generalization capability of a neural captioning system given only a few keywords. 

In our method, we do not require a target keyword $K_j,~j \in [M]$ to appear at a particular position. Instead, we want a loss function that allows $K_j$ to become the top-1 prediction (plus a confidence margin $\epsilon$) at any position. 
Therefore, we propose to use the minimum of the hinge-like loss terms over all $t \in [N]$ as an indication of $K_j$ appearing at any position as the top-1 prediction, leading to the following loss function:
\begin{equation}
\label{eq:keyword_loss1}
\!\!\!\!\text{loss}_{K, \text{logits}} = \!\!\!\sum_{j=1}^{M} \min_{t \in [N]}\{ \max \{ -\epsilon,\! \max_{k\neq K_j}\{z^{(k)}_t\}-z_t^{(K_j)}\}\}.
\end{equation}

We note that the loss functions in~\eqref{eq:rnn} and~\eqref{eq:softmaxloss} 
require an input $S^\prime_{t-1}$ to predict $z_t$ for each $t \in \{2,\ldots,N\}$. For the targeted caption method, we use the targeted caption $S$ as the input of RNN. 
In contrast, for the targeted keyword method we no longer know the exact targeted sentence, but only require the presence of specified keywords in the final caption.
To bridge the gap, we use the originally inferred caption $S^0 = (S^0_1, \cdots, S^0_N)$ from the benign image as the initial input to RNN.
Specifically, after minimizing~\eqref{eq:keyword_loss1} for $T$ iterations, we run inference on $I+\delta$ and set the RNN's input $S^{1}$ as its current top-1 prediction, and continue this process. With this iterative optimization process, the desired keywords are expected to gradually appear in top-1 prediction.

Another challenge arises in targeted keyword method is the problem of ``keyword collision''. When the number of keywords $M \geq 2$, more than one keywords may have large values of $\max_{k\neq K_j}\{z^{(k)}_t\}-z_t^{(K_j)}$ at a same position $t$. For example, if \texttt{dog} and \texttt{cat} are top-2 predictions for the second word in a caption, the caption can either start with ``A dog ...'' or ``A cat ...''. In this case, despite~the loss \eqref{eq:keyword_loss1} being very small, a caption with both \texttt{dog} and \texttt{cat} can hardly be generated, since only one word is allowed to appear at the same position. To alleviate this problem, we define a gate function $g_{t,j}(x)$ which masks off all the other keywords when a keyword becomes top-$1$ at position $t$:
\begin{equation}
g_{t,j}(x)=\left\{
\begin{array}{ll}
\!A,\ \text{if~}\arg\max_{i \in \mathcal{V}} z_t^{(i)} \in \mathcal{K} \setminus \{K_j\} \\
x,\ \text{otherwise},\nonumber
\end{array}
\right.
\end{equation}
where A is a predefined value that is significantly larger than common logits values. Then~\eqref{eq:keyword_loss1} becomes:
\begin{align}
\label{eq:keyword_loss2}
&\sum_{j=1}^{M} \min_{t \in [N]}\{ g_{t,j}(\max \{ -\epsilon, \max_{k\neq K_j}\{z^{(k)}_t\}-z_t^{(K_j)} \}) \}.
\end{align}
The log-prob loss for targeted keyword method is discussed in the Supplementary Material.

\section{Experiments}
\subsection{Experimental Setup and Algorithms}
We performed extensive experiments to test the effectiveness of our Show-and-Fool algorithm and study the robustness of image captioning systems under different problem settings. In our experiments\footnote{{Our source code is available at: \url{https://github.com/huanzhang12/ImageCaptioningAttack}}}, we use the pre-trained TensorFlow implementation\footnote{\url{https://github.com/tensorflow/models/tree/master/research/im2txt}} 
of Show-and-Tell~\cite{DBLP:conf/cvpr/VinyalsTBE15} with Inception-v3 as the CNN for visual feature extraction. Our 
testbed is Microsoft COCO~\cite{lin2014microsoft} (MSCOCO) data set. Although some more recent neural image captioning systems can achieve better performance than Show-and-Tell, they share a similar 
framework that uses CNN for feature extraction and RNN for caption generation, and Show-and-Tell is the vanilla version of this CNN+RNN architecture. Indeed, we find that the adversarial examples on Show-and-Tell are transferable to other image captioning models such as Show-Attend-and-Tell~\cite{DBLP:conf/icml/XuBKCCSZB15} and NeuralTalk2\footnote{\url{https://github.com/karpathy/neuraltalk2}}, 
suggesting that the attention mechanism and the choice of CNN and RNN architectures do not significantly affect the robustness. We also note that since Show-and-Fool is the first work on crafting adversarial examples for neural image captioning, to the best of our knowledge, there is no other method for comparison.

We use ADAM to minimize our loss functions and set the learning rate to 0.005. The number of iterations is set to $1,000$. All the experiments are performed on a single Nvidia GTX 1080 Ti GPU. For targeted caption and targeted keyword methods, we perform a binary search for 5 times to find the best $c$: initially $c=1$, and $c$ will be increased by $10$ times until a successful adversarial example is found. Then, we choose a new $c$ to be the average of the largest $c$ where an adversarial example can be found and the smallest $c$ where an adversarial example cannot be found. We fix $\epsilon = 1$ except for transferability experiments. For each experiment, we randomly select 1,000 images from the MSCOCO validation set. 
We use BLEU-1~\cite{papineni2002bleu}, BLEU-2, BLEU-3, BLEU-4, ROUGE~\cite{lin2004rouge} and METEOR~\cite{lavie2005meteor} scores to evaluate the correlations between the inferred captions and the targeted captions. These scores are widely used in NLP community and are adopted by image captioning systems for quality assessment. Throughout this  section, we use the \textit{logits loss} \eqref{eq:logits loss}\eqref{eq:keyword_loss2}. The results of using the \textit{log-prob loss}  \eqref{eq:softmaxloss} are similar and are reported in the supplementary material. 

\subsection{Targeted Caption Results}
\label{sec:exp_targeted}
\begin{table}[t]
\small
\centering
\caption{Summary of targeted caption method (Section \ref{sec:exact}) and targeted keyword method (Section \ref{sec:keyword}) using logits loss. The $\ell_2$ distortion of adversarial noise $\|\delta\|_2$ is averaged over successful adversarial examples. For comparison, we also include CNN based attack methods (Section~\ref{attack_CNN}).}
\begin{tabular*}{0.9\columnwidth}{c|cc} \toprule
Experiments&Success Rate&Avg. $\|\delta\|_2$ \\ \specialrule{0.85pt}{0pt}{0pt}
targeted caption &95.8\%&2.213\\ \hline
1-keyword&97.1\% &1.589\\\hline
2-keyword&97.5\% &2.363\\\hline
3-keyword&96.0\% &2.626\\\hline
C\&W on CNN&22.4\% &2.870\\\hline
I-FGSM on CNN&34.5\% &15.596\\\bottomrule
\end{tabular*}
\label{tab:keywords result}
\end{table}

Unlike the image classification task where all possible labels are predefined, the space of possible captions in a captioning system is almost infinite. However, the captioning system is only able to output relevant captions learned from the training set. For instance, the captioning model cannot generate a passive-voice sentence if the model was never trained on such sentences. 
Therefore, we need to ensure that the targeted caption lies in the space where the captioning system can possibly generate. 
To address this issue, we use the generated caption of a randomly selected image (other than the image under investigation) from MSCOCO validation set as the targeted caption $S$. The use of a generated caption as the targeted caption excludes the effect of out-of-domain captioning, and ensures that the target caption is within the output space of the captioning network.

\begin{table}[t]
\small
\centering
\caption{Statistics of the 4.2\% failed adversarial  examples using the targeted caption method and logits loss (\ref{eq:logits loss}). All correlation scores are computed using the top-5 inferred captions of an adversarial image and the targeted caption (higher score means better targeted attack performance).}
\begin{tabular*}{\columnwidth}{c|ccccc} \toprule
$c$&1&10&$10^2$&$10^3$&$10^4$\\\hline
$\ell_2$ Distortion&1.726&3.400&7.690&16.03&23.31\\\hline
BLEU-1&.567&.725&.679&.701&.723\\\hline
BLEU-2&.420&.614&.559&.585&.616\\\hline
BLEU-3&.320&.509&.445&.484&.514\\\hline
BLEU-4&.252&.415&.361&.402&.417\\\hline
ROUGE&.502&.664&.629&.638&.672\\\hline
METEOR&.258&.407&.375&.403&.399\\\bottomrule
\end{tabular*}
\label{tab:logits targeted fail avg score}
\end{table}

Here we use the logits loss \eqref{eq:logits loss} plus a $\ell_2$ distortion term (as in \eqref{eq:general_loss}) as our objective function. A successful adversarial example is found if the inferred caption after adding the adversarial perturbation $\delta$ is \textit{exactly the same} as the targeted caption. In our setting, 1,000 ADAM iterations take about 38 seconds for one image.
The overall success rate and average distortion of adversarial perturbation $\delta$ are shown in Table~\ref{tab:keywords result}. Among all the tested images, our method attains 95.8\% attack success rate. 
Moreover, our adversarial examples have small $\ell_2$ distortions and are visually identical to the original images, as displayed in Figure~\ref{Fig_show_fool}. We also examine the failed adversarial examples and summarize their statistics in Table \ref{tab:logits targeted fail avg score}. We find that their generated captions, albeit not entirely identical to the targeted caption, are in fact highly correlated to the desired one. 
Overall, the high success rate and low $\ell_2$ distortion of adversarial examples clearly show that Show-and-Tell is not robust to targeted adversarial perturbations.

\subsection{Targeted Keyword Results}\label{sec:exp_keyword}
\begin{figure}[!t]
    \centering
    \includegraphics[width=0.95\linewidth]{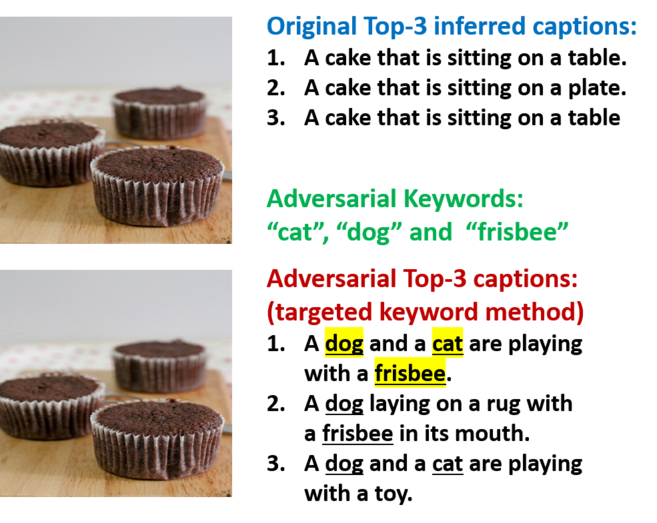}
    \caption{An adversarial example ($\|\delta\|_2 = 1.284$) of an cake image crafted by the Show-and-Fool targeted keyword method with three keywords - ``dog'', ``cat'' and ``frisbee''.}
    \label{Fig_cake}
\end{figure}

In this task, we use~(\ref{eq:keyword_loss2}) as our loss function, and choose the number of keywords $M=\{1, 2, 3\}$. We run an inference step on $I+\delta$ every $T=5$ iterations, and use the top-1 caption as the input of RNN/LSTMs. Similar to Section~\ref{sec:exp_targeted}, for each image the targeted keywords are selected from the caption generated by a randomly selected validation set image. To exclude common words like ``a'', ``the'', ``and'', we look up each word in the targeted sentence and only select nouns, verbs, adjectives or adverbs. We say an adversarial image is successful when its caption contains \textit{all} specified keywords. The overall success rate and average distortion are shown in Table~\ref{tab:keywords result}.
When compared to the targeted caption method, targeted keyword method achieves an even higher success rate (at least 96\% for 3-keyword case and at least 97\% for 1-keyword and 2-keyword cases). 
Figure \ref{Fig_cake} shows an adversarial example crafted from our targeted keyword method with three keywords - ``dog'', ``cat'' and ``frisbee''. Using Show-and-Fool, the top-1 caption of a cake image becomes ``A dog and a cat are playing with a frisbee'' while the adversarial image remains visually indistinguishable to the original one. When $M = 2$ and $3$, even if we cannot find an adversarial image yielding all specified keywords, we might end up with a caption that contains some of the keywords (partial success).  For example, when $M=3$, Table~\ref{tab:3 keyword fail avg score logits} shows the number of keywords appeared in the captions ($M'$) for those \textit{failed} examples (not all 3 targeted keywords are found). These results clearly show that the $4\%$ failed examples are still partially successful: the generated captions contain about 1.5 targeted keywords on average. 

\begin{table}[ht]
\small
\centering
\caption{Percentage of partial success with different $c$ in the 4.0\% failed images that do not contain all the $3$ targeted keywords. }
\begin{tabular*}{0.9\columnwidth}{c|cccc} \toprule
$c$&Avg. $\|\delta\|_2$& $M' \geq 1$ & $M'=2$ & Avg. $M'$\\\specialrule{0.85pt}{0pt}{0pt}
$1$&2.49&72.4\%&34.5\%&1.07\\\hline
$10$&5.40&82.7\%&37.9\%&1.21\\\hline
$10^2$&12.95&93.1\%&58.6\%&1.52\\\hline
$10^3$&24.77&96.5\%&51.7\%&1.48\\\hline
$10^4$&29.37&100.0\%&58.6\%&1.59\\\bottomrule
\end{tabular*}
\label{tab:3 keyword fail avg score logits}
\end{table}

\subsection{Transferability of Adversarial Examples}
\label{subsec_transfer}
\input{table_transferability}
It has been shown that in image classification tasks, adversarial
examples found for one machine learning model may also be effective against another model, even if the two models have different architectures~\cite{papernot2016transferability,liu2016delving}.
However, unlike image classification where correct labels are made explicit, two different image captioning systems may generate quite different, yet semantically similar, captions for the same benign image.  In image captioning, we say an adversarial example is \textit{transferable} when the adversarial image found on model $A$ with a target sentence $S_A$ can generate a similar (rather than exact) sentence $S_B$ on model $B$. 

In our setting, model $A$ is Show-and-Tell, and we choose Show-Attend-and-Tell~\cite{DBLP:conf/icml/XuBKCCSZB15} as model $B$. The major differences between Show-and-Tell and Show-Attend-and-Tell are the addition of attention units in LSTM network for caption generation, and the use of last convolutional layer (rather than the last fully-connected layer) feature maps for feature extraction. We use Inception-v3 as the CNN architecture for both models and train them on the MSCOCO 2014 data set. 
However, their CNN parameters are different due to the fine-tuning process.

\begin{figure}[!t]
    \centering
    \includegraphics[width=0.95\linewidth]{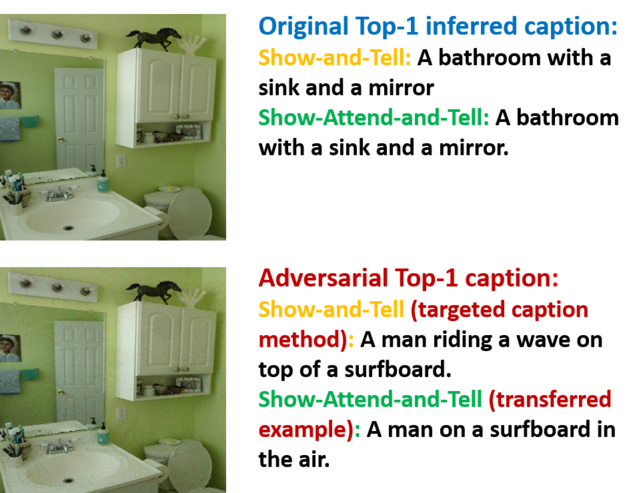}
    \caption{A highly transferable adversarial example ($\|\delta\|_2 = 15.226$) crafted by Show-and-Tell targeted caption method, transfers to  Show-Attend-and-Tell, yielding similar adversarial captions.
}
    \label{Fig_bathroom}
\end{figure}

To investigate the transferability of adversarial examples in image captioning, we first use the targeted caption method to find adversarial examples for 1,000 images in model $A$ with different $c$ and $\epsilon$, and then transfer successful adversarial examples (which generate the exact target captions on model $A$) to model $B$. The generated captions by model $B$ are recorded for transferability analysis. The  transferability of adversarial examples depends on two factors: the intrinsic difference between two models even when the same benign image is used as the input, i.e., \textit{model mismatch}, and the transferability of adversarial perturbations.

To measure the mismatch between Show-and-Tell and Show-Attend-and-Tell, we generate captions of the same set of 1,000 original images from both models, and report their mutual BLEU, ROUGE and METEOR scores in Table~\ref{tab:transfer} under the \textbf{mis} column.
To evaluate the effectiveness of transferred adversarial examples, we measure the scores for two set of captions: (i) the captions of original images and the captions of transferred adversarial images, both generated by Show-Attend-and-Tell (shown under column \textbf{ori} in Table~\ref{tab:transfer}); and (ii) the targeted captions for generating adversarial examples on Show-and-Tell, and the captions of the  transferred  adversarial image on Show-Attend-and-Tell (shown under column \textbf{tgt} in Table~\ref{tab:transfer}). Small values of \textbf{ori} suggest that the adversarial images on Show-Attend-and-Tell generate significantly different captions from original images' captions. Large values of \textbf{tgt} suggest that the adversarial images on Show-Attend-and-Tell generate similar adversarial captions as on the Show-and-Tell model. We find that increasing $c$ or $\epsilon$ helps to enhance transferability at the cost of larger (but still acceptable) distortion. When $C=1,000$ and $\epsilon=10$, Show-and-Fool achieves the best transferability results: \textbf{tgt} is close to \textbf{mis}, indicating that the discrepancy between adversarial captions on the two models is mostly bounded by the intrinsic model mismatch rather than the transferability of adversarial perturbations, and implying that the adversarial perturbations are easily transferable. 
In addition, the adversarial examples generated by our method can also fool NeuralTalk2. When $c=10^4, \epsilon=10$, the average $\ell_2$ distortion, BLEU-4 and METEOR scores between the original and transferred adversarial captions are 38.01, 0.440 and 0.473, respectively. 
The high transferability of adversarial examples crafted by Show-and-Fool also indicates the problem of common robustness leakage between different neural image captioning models.

\subsection{Attacking Image Captioning v.s. Attacking Image Classification}
\label{attack_CNN}
In this section we show that attacking image captioning models is inherently more challenging than attacking image classification models. In the classification task, a targeted attack usually becomes harder when the number of labels increases, since an attack method needs to change the classification prediction to a specific label over all the possible labels. In the targeted attack on image captioning, if we treat each caption as a label, we need to change the original label to a specific one over an almost infinite number of possible labels, corresponding to a nearly zero volume in the search space. This constraint forces us to develop non-trivial methods that are significantly different from the ones designed for attacking image classification models. 

To verify that the two tasks are inherently different, we conducted additional experiments on attacking \textit{only} the CNN module using two state-of-the-art image classification attacks on ImageNet dataset. Our experiment setup is as follows. Each selected ImageNet image has a label corresponding to a WordNet synset ID. 
We randomly selected 800 images from ImageNet dataset such that their synsets have at least one word in common with Show-and-Tell’s vocabulary, while ensuring the Inception-v3 CNN (Show-and-Tell’s CNN) classify them correctly. Then, we perform Iterative Fast Gradient Sign Method (I-FGSM) \cite{kurakin2016adversarial_ICLR} and Carlini and Wagner’s (C\&W) attack \cite{carlini2017towards} on these images. The attack target labels are randomly chosen and their synsets also have at least one word in common with Show-and-Tell’s vocabulary. Both I-FGSM and C\&W achieve 100\% targeted attack success rate on the Inception-v3 CNN. These adversarial examples were further employed to attack Show-and-Tell model. An attack is considered successful if \textit{any} word in the targeted label’s synset or its hypernyms up to 5 levels is presented in the resulting caption. For example, for the chain of hypernyms `broccoli'$\Rightarrow$`cruciferous vegetable'$\Rightarrow$`vegetable, veggie, veg'$\Rightarrow$`produce, green goods, green groceries, garden truck'$\Rightarrow$`food, solid food', we include `broccoli',`cruciferous',`vegetable',`veggie' and all other following words. Note that this criterion of success is much weaker than the criterion we use in the targeted caption method, since a caption with the targeted image's hypernyms does not necessarily leads to similar meaning of the targeted image's captions. To achieve higher attack success rates, we allow relatively larger distortions and set $\epsilon_\infty=0.3$ (maximum $\ell_\infty$ distortion) in I-FGSM and $\kappa=10$, $C=100$ in C\&W. However, as shown in Table \ref{tab:keywords result}, the attack success rates are only 34.5\% for I-FGSM and 22.4\% for C\&W, respectively, which are much lower than the success rates of our methods despite larger distortions. This result further confirms that performing targeted attacks on neural image captioning requires a careful design (as proposed in this paper), and attacking image captioning systems is not a trivial extension to attacking image classifiers.

\section{Conclusion}


In this paper, we proposed a novel algorithm, {\bf Show-and-Fool},
for crafting adversarial examples and providing robustness evaluation of neural image captioning. Our extensive experiments show that the proposed targeted caption and keyword methods yield high attack success rates while the adversarial perturbations are still imperceptible to human eyes. 
We further demonstrate that Show-and-Fool can generate highly transferable adversarial examples. 
The high-quality and transferable adversarial examples in neural image captioning crafted by Show-and-Fool highlight the inconsistency in visual language grounding between humans and machines, suggesting
a possible weakness of current machine vision and perception machinery. We also show that attacking neural image captioning systems are inherently different from attacking CNN-based image classifiers.

Our method stands out from the well-studied adversarial learning on image classifiers and CNN models. To the best of our knowledge, this is the very first work on crafting adversarial examples for neural image captioning systems. Indeed, our Show-and-Fool algorithm\footnotemark[1] can be easily extended to other applications with RNN or CNN+RNN architectures. We believe this paper provides potential means to evaluate and possibly improve the robustness (for example, by adversarial training or data augmentation) of a wide range of visual language grounding and other NLP models.




{
\bibliographystyle{acl_natbib}
\bibliography{paperbib}
}

\clearpage
\section*{\Large{Supplementary Material}}


\input{supp}

\end{document}

%% file: table_transferability.tex
\begin{table*}[htbp]
\centering
\caption{Transferability of adversarial examples from Show-and-Tell to Show-Attend-and-Tell, using different $\epsilon$ and $c$. \textbf{ori} indicates the scores between the generated captions of the \textit{original} images and the transferred adversarial images on Show-Attend-and-Tell. \textbf{tgt} indicates the scores between the \textit{targeted} captions on Show-and-Tell and the generated captions of transferred adversarial images on Show-Attend-and-Tell. A smaller \textbf{ori} or a larger \textbf{tgt} value indicates better transferability. \textbf{mis} measures the differences between captions generated by the two models given the same benign image (\textit{model mismatch}). When $C=1000, \epsilon=10$,  \textbf{tgt} is close to \textbf{mis}, indicating the discrepancy between adversarial captions on the two models is mostly bounded by model mismatch, and the adversarial perturbation is highly transferable.}
\label{tab:transfer}
\scalebox{0.7}{
\begin{tabular}{cll|ll|ll|ll|ll|ll|ll|ll|ll|ll}
\cline{2-20}
                                          & \multicolumn{6}{c|}{$\epsilon = 1$}                                                 & \multicolumn{6}{c|}{$\epsilon = 5$}                                                & \multicolumn{6}{c|}{$\epsilon = 10$}                                                &         \\ \cline{2-20} 
                                          &\multicolumn{2}{c|}{C=10} & \multicolumn{2}{c|}{C=100} & \multicolumn{2}{c|}{C=1000} &\multicolumn{2}{c|}{C=10} & \multicolumn{2}{c|}{C=100} & \multicolumn{2}{c|}{C=1000}&\multicolumn{2}{c|}{C=10}  & \multicolumn{2}{c|}{C=100} & \multicolumn{2}{c|}{C=1000}&         \\ \cline{2-20} 
                                          & \bf{ori}  & \bf{tgt}     & \bf{ori}    & \bf{tgt}     & \bf{ori}    & \bf{tgt}      & \bf{ori}  & \bf{tgt}     & \bf{ori}    & \bf{tgt}     & \bf{ori}    & \bf{tgt}     & \bf{ori}  & \bf{tgt}      & \bf{ori}    & \bf{tgt}     & \bf{ori}    & \bf{tgt}     & \bf{mis}\\ \hline
\multicolumn{1}{l|}{\!\!BLEU-1}           &  .474     &  .395        &  .384       &  .462        &  .347       &  .484         &  .441     &  .429        &  .368       &  .488        & \bf{.337}   &  .527        &  .431     &  .421         &  .360       &  .485        &    {.339}   & \bf{.534}    & .649    \\
\multicolumn{1}{l|}{\!\!BLEU-2}           &  .337     &  .236        &  .230       &  .331        &  .186       &  .342         &  .300     &  .271        &  .212       &  .343        &  .175       &  .389        &  .287     &  .266         &  .204       &  .342        & \bf{.174}   & \bf{.398}    & .521    \\
\multicolumn{1}{l|}{\!\!BLEU-3}           &  .256     &  .154        &  .151       &  .224        &  .114       &  .254         &  .220     &  .184        &  .135       &  .254        &  .103       &  .299        &  .210     &  .185         &  .131       &  .254        & \bf{.102}   & \bf{.307}    & .424    \\
\multicolumn{1}{l|}{\!\!BLEU-4}           &  .203     &  .109        &  .107       &  .172        &  .077       &  .198         &  .170     &  .134        &  .093       &  .197        &  .068       &  .240        &  .162     &  .138         &  .094       &  .197        & \bf{.066}   & \bf{.245}    & .352    \\ \hline
\multicolumn{1}{l|}{\!\!ROUGE}            &  .463     &  .371        &  .374       &  .438        &  .336       &  .465         &  .429     &  .402        &  .359       &  .464        &  .329       &  .502        &  .421     &  .398         &  .351       &  .463        & \bf{.328}   & \bf{.507}    & .604    \\ \hline
\multicolumn{1}{l|}{\!\!METEOR}           &  .201     &  .138        &  .139       &  .180        &  .118       &  .201         &  .177     &  .157        &  .131       &  .199        &  .110       &  .228        &  .172     &  .157         &  .127       &  .202        & \bf{.110}   & \bf{.232}    & .300    \\ \hline
\multicolumn{1}{l|}{\!\!$\|\delta\|_2$}   &\multicolumn{2}{c|}{3.268}& \multicolumn{2}{c|}{4.299} & \multicolumn{2}{c|}{4.474}  &\multicolumn{2}{c|}{7.756}& \multicolumn{2}{c|}{10.487}& \multicolumn{2}{c|}{10.952}&\multicolumn{2}{c|}{15.757}& \multicolumn{2}{c|}{21.696}& \multicolumn{2}{c|}{21.778}&         \\ \hline
\end{tabular}
}
\end{table*}

%% file: supp.tex
\section{Related Work on Adversarial Attacks to CNN-based Image Classifiers}
Despite the remarkable progress, CNNs have been shown to be vulnerable to adversarial examples \cite{szegedy2013intriguing,goodfellow2014explaining,carlini2017towards}. In image classification, an adversarial example is an image that is visually indistinguishable to the original image but can cause a CNN model to misclassify. With different objectives, adversarial attacks can be divided into two categories, i.e., untargeted attack and targeted attack. In the literature, a successful untargeted attack refers to finding an adversarial example that is close to the original example but yields different class prediction. For targeted attack, a target class is specified and the adversarial example is considered successful when the predicted class matches the target class. Surprisingly, adversarial examples can also be crafted even when the parameters of target CNN model are unknown to an attacker \cite{liu2016delving,CPY17_zoo}. In addition, adversarial examples crafted from one image classification model can be made transferable to other models \cite{liu2016delving,papernot2016transferability}, and there exists a universal adversarial perturbation that can lead to misclassification of natural images with high probability \cite{moosavi2016universal}.

 Without loss of generality, there are two factors contributing to crafting adversarial examples in image classification: (i) a distortion metric between the original and adversarial examples that regularizes visual similarity. Popular choices are the $L_\infty$, $L_2$ and  $L_1$ distortions \cite{kurakin2016adversarial_ICLR,carlini2017towards,chen2017ead}; and (ii) an attack loss function accounting for the success of adversarial examples.
For finding adversarial examples in neural image captioning, 
while the distortion metric can be identical, the attack loss function used in image classification is invalid, since the number of possible captions easily outnumbers the number of image classes, and captions with similar meaning should not be considered as different classes. One of our major contributions is to design novel attacking loss functions to handle the CNN+RNN architectures in neural image captioning tasks.

\section{More Adversarial Examples with Logits Loss}
\begin{figure}[h]
	\centering
	\includegraphics[width=0.99\columnwidth]{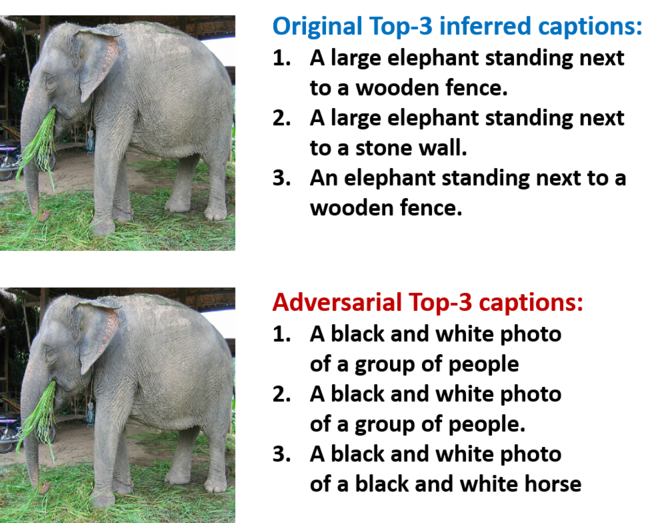}
	\caption{Adversarial example ($\|\delta\|_2 = 2.977$) of an elephant image crafted by the Show-and-Fool targeted caption method with the target caption ``A black and white photo of a group of people''.}
	\label{Fig_elephant}
\end{figure}

\begin{figure}[htbp]
	\centering
	\includegraphics[width=0.99\columnwidth]{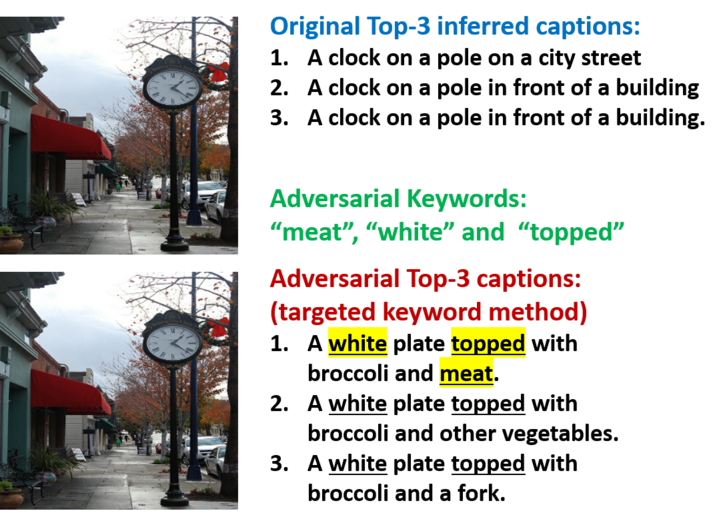}
	\caption{Adversarial example ($\|\delta\|_2 = 2.979$) of an clock image crafted by the Show-and-Fool targeted keyword method with three keywords:  ``meat'', ``white'' and ``topped''.}
	\label{Fig_clock}
\end{figure}

\begin{figure}[t]
	\centering
	\includegraphics[width=1.02\columnwidth]{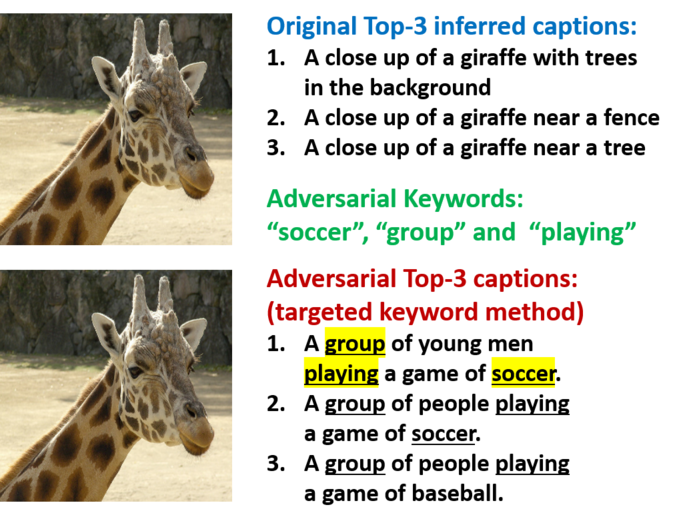}
	\caption{Adversarial example ($\|\delta\|_2 = 1.188$) of a giraffe image crafted by the Show-and-Fool targeted keyword method with three keywords:  ``soccer'', ``group'' and ``playing''.}
	\label{Fig_giraffe}
\end{figure}

\begin{figure}[htbp]
	\centering
	\includegraphics[width=1.02\columnwidth]{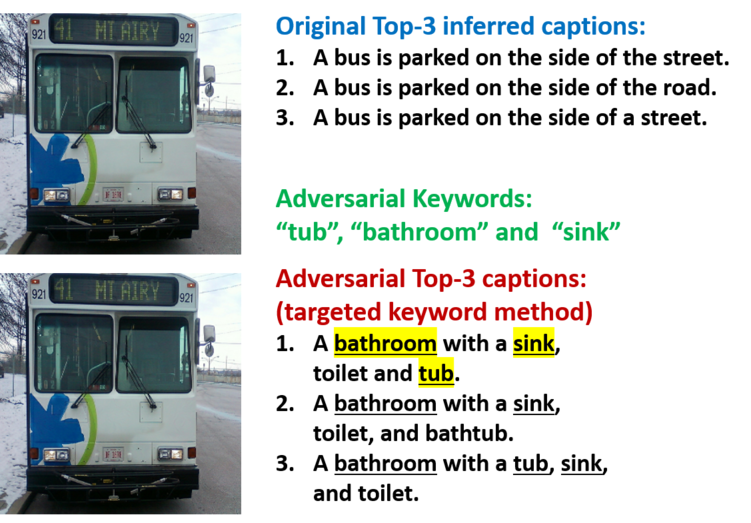}
	\caption{Adversarial example ($\|\delta\|_2 = 1.178$) of a bus image crafted by the Show-and-Fool targeted keyword method with three keywords:  ``tub'', ``bathroom'' and ``sink''.}
	\label{Fig_bus}
\end{figure}

Figure \ref{Fig_elephant} shows another successful example with targeted caption method. 
Figures \ref{Fig_clock}, \ref{Fig_giraffe} and \ref{Fig_bus} show three adversarial examples generated by the proposed 3-keyword method. The adversarial examples generated by our methods have small $L_2$ distortions and are visually indistinguishable from the original images. One advantage of using logits losses is that it helps to bypass defensive distillation by overcoming the gradient vanishing problem. To see this, the partial derivative of the softmax function 
$$
p^{(j)}=\exp(z^{(j)})/\sum_{i\in\mathcal{V}}\exp(z^{(i)}),
$$
is given by
\begin{equation}
\frac{\partial p^{(j)}}{\partial z^{(j)}}=p^{(j)}(1-p^{(j)}),
\end{equation}
which vanishes as $p^{(j)}\to 0$ or $p^{(j)}\to 1.$ The defensive distillation method [30] 
uses a large distillation temperature in the training process and removes it in the inference process. This makes the inference probability $p^{(j)}$ close to 0 or 1, thus leads to a vanished gradient problem. However, by using the proposed logits loss (\ref{eq:logits loss}), before the word at position $t$ in target sentence $S$ reaches top-1 probability, we have
\begin{equation}
\frac{\partial}{\partial z_t^{(S_t)}}\text{loss}_{S,\text{logits}}(I+\delta)=-1.
\end{equation}
It is evident that the gradient  (with regard to $z_t^{(S_t)}$) becomes a constant now, since it equals to $-1$ when $z_t^{(S_t)} < \max_{k\neq S_t}\{z^{(k)}_t\} + \epsilon,$ and 0 otherwise.

\begin{figure}[t]
	\centering
	\includegraphics[width=0.99\columnwidth]{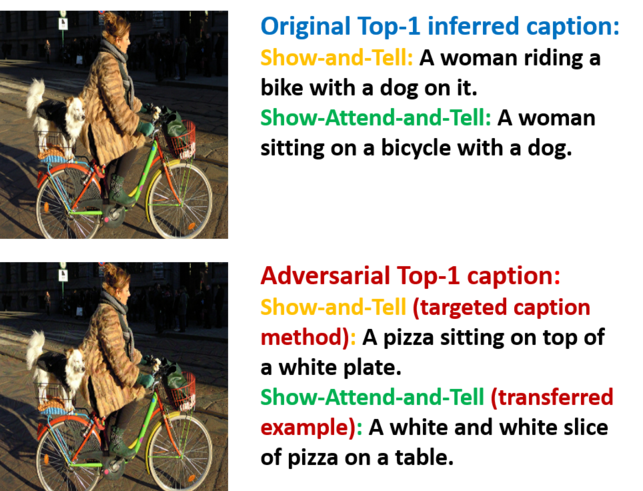}
	\caption{A highly transferable adversarial example of a biking image ($\|\delta\|_2 = 12.391$) crafted from Show-and-Tell using the targeted caption method and then transfers to Show-Attend-and-Tell, yielding similar adversarial captions.
}
	\label{Fig_womanbike}
\end{figure}

\begin{figure}[htbp]
	\centering
	\includegraphics[width=0.99\columnwidth]{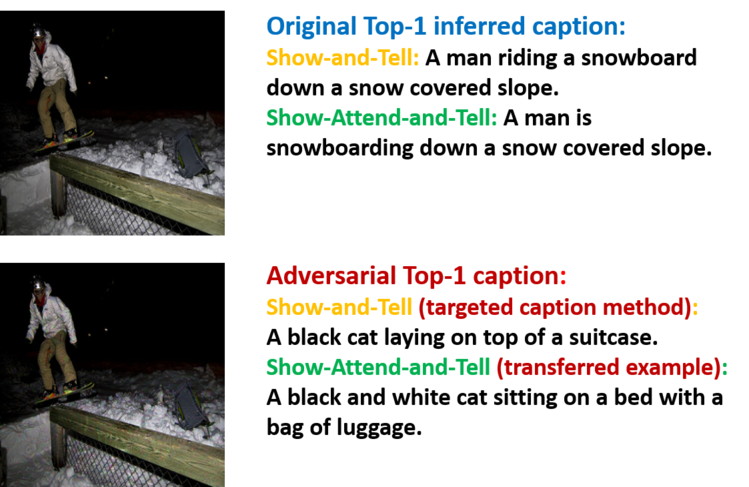}
	\caption{A highly transferable adversarial example of a snowboarding image ($\|\delta\|_2 = 14.320$) crafted from Show-and-Tell using the targeted caption method and then transfers to Show-Attend-and-Tell, yielding similar adversarial captions.
}
	\label{Fig_snowboard}
\end{figure}

\begin{figure}[ht]
	\centering
\includegraphics[width=1.02\columnwidth]{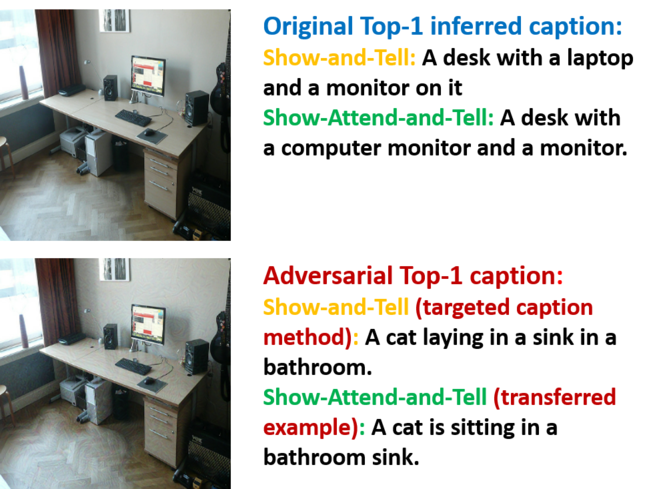}
	\caption{A highly transferable adversarial example of a desk image ($\|\delta\|_2 = 12.810$) crafted from Show-and-Tell using the targeted caption method and then transfers to  Show-Attend-and-Tell, yielding similar adversarial captions.
}
	\label{Fig_desk}
\end{figure}
\section{Targeted Caption Results with Log Probability Loss}
\begin{table}[htbp]
\small
\centering
\caption{Summary of targeted caption method and targeted keyword method using \textit{log-prob} loss. The $L_2$ distortion $\|\delta\|_2$ is averaged over successful adversarial examples.}
\begin{tabular*}{0.8\columnwidth}{c|cc} \toprule
Experiments&Success Rate&Avg. $\|\delta\|_2$ \\ \specialrule{0.85pt}{0pt}{0pt}
targeted caption &95.4\%&1.858\\ \hline
1-keyword&99.2\% &1.311\\\hline
2-keyword&96.9\% &2.023\\\hline
3-keyword&95.7\% &2.120\\\bottomrule
\end{tabular*}
\label{tab:logprob keywords result}
\end{table}

\begin{table}[htbp]
\small
\centering
\caption{Statistics of the 4.6\% failed adversarial  examples using the targeted caption method and \textit{log-prob} loss (\ref{eq:softmaxloss}). All correlation scores are computed using the top-5 inferred captions of an adversarial image and the targeted caption (a higher score indicates a better targeted attack performance).}
\begin{tabular*}{\columnwidth}{c|ccccc} \toprule
$c$&1&10&$10^2$&$10^3$&$10^4$\\\hline
$L_2$ Distortion&1.503&2.637&5.085&11.15&19.69\\\hline
BLEU-1&.650&.792&.775&.802&.800\\\hline
BLEU-2&.521&.690&.671&.711&.701\\\hline
BLEU-3&.416&.595&.564&.622&.611\\\hline
BLEU-4&.354&.515&.485&.542&.531\\\hline
ROUGE&.616&.764&.746&.776&.772\\\hline
METEOR&.362&.493&.469&.511&.498\\\bottomrule
\end{tabular*}
\label{tab:log prob targeted fail avg score}
\end{table}

In this experiment, we use the log probability loss \eqref{eq:softmaxloss} plus a $L_2$ distortion term (as in \eqref{eq:general_loss}) as our objective function. Similar to the previous experiments, a successful adversarial example is found if the inferred caption after adding the adversarial perturbation $\delta$ \textit{exactly matches} the targeted caption. The overall success rate and average distortion of adversarial perturbation $\delta$ are shown in Table~\ref{tab:logprob keywords result}. Among all the tested images, our log-prob loss attains 95.4\% success rate, which is about the same as using logits loss. Besides, similar to using logits loss, the adversarial examples generated by using log-prob loss also yield small $L_2$ distortions. In Table \ref{tab:log prob targeted fail avg score}, we summarize the statistics of the failed adversarial examples. It shows that their generated captions, though not entirely identical to the targeted caption, are also highly relevant to the target captions. 

In our experiments, log probability loss exhibits a similar performance as the logits loss, as our target model is undefended and the gradient vanishing problem of softmax is not significant. However, when evaluating the robustness of a general image captioning model, it is recommended to use the logits loss as it does not suffer from potentially vanished gradients and can reveal the intrinsic robustness of the model.

\section{Targeted Keyword Results with Log Probability Loss}\label{supp keyword}

Similar to the logits loss, the log-prob loss does not require a particular position for the target keywords $K_j, j \in [M]$. Instead, it encourages $K_j$ to become the top-1 prediction at its most probable position: 
\begin{equation}
\label{eq:keyword_logprob1}
\!\!\text{loss}_{K, \text{log-prob}} = -\sum_{j=1}^{M}\log(\max_{t \in [N]}\{p_t^{(i)}\}).
\end{equation}
To tackle the ``keyword collision'' problem, we also employ a gate function $g'_{t,j}$ to avoid the keywords appearing at the positions where the most probable word is already a keyword:
\begin{equation}
g'_{t,j}(x)=\left\{
\begin{array}{ll}
\!0,\ \text{if~}\arg\max_{i \in \mathcal{V}} p_t^{(i)} \in \mathcal{K} \setminus \{K_j\} \\
x,\ \text{otherwise}\nonumber
\end{array}
\right.
\end{equation}
The loss function (\ref{eq:keyword_logprob1}) then becomes:
\begin{equation}
\label{eq:keyword_logprob2}
\text{loss}_{K^\prime, \text{log-prob}} = -\sum_{j=1}^{M} \log(\max_{t \in [N]}\{ g'_{t,j}(p_t^{(i)})\}).
\end{equation}
In our methods, the initial input is the originally inferred caption $S^0$ from the benign image, and after minimizing~\eqref{eq:keyword_logprob2} for $T$ iterations, we run inference on $I+\delta$ and set the RNN's input $S^{1}$ as its current top-1 prediction, and repeat this procedure until all the targeted keywords are found or the maximum number of iterations is met. With this iterative optimization process, the probabilities of the desired keywords gradually increase, and finally become the top-1 predictions. 

\begin{table}[htbp]
\small
\centering
\caption{Percentage of partial success using \textit{log-prob} loss with different $c$ in the 4.3\% failed images that do not contain all the $3$ targeted keywords.}
\begin{tabular*}{0.9\columnwidth}{c|cccc} \toprule
$c$&Avg. $\|\delta\|_2$& $M' \geq 1$ & $M' = 2$ & Avg. $M'$\\\specialrule{0.85pt}{0pt}{0pt}
$1$&2.22&69.7\%&27.3\%&0.97\\\hline
$10$&5.03&87.9\%&57.6\%&1.45\\\hline
$10^2$&10.98&93.9\%&63.6\%&1.58\\\hline
$10^3$&18.52&93.9\%&57.6\%&1.52\\\hline
$10^4$&26.04&90.9\%&60.6\%&1.52\\\bottomrule
\end{tabular*}
\label{tab:3 keyword fail avg score logprob}
\end{table}

The overall success rate and average distortion are shown in Table~\ref{tab:logprob keywords result}. Table~\ref{tab:3 keyword fail avg score logprob} summarizes the number of keywords ($M'$) appeared in the captions for those \textit{failed} examples when $M=3$, i.e., the examples that not all the 3 targeted keywords are found. They account only $4.3\%$ of all the tested images. Table~\ref{tab:3 keyword fail avg score logprob} clearly shows that when $c$ is properly chosen, more than $90\%$ of the failed examples contain at least 1 targeted keyword, and more than $60\%$ of the failed examples contain 2 targeted keywords. This result verifies that even the failed examples are reasonably good attacks.

\vspace{-0.05cm}
\section{Transferability of Adversarial Examples with Log Probability Loss}
Similar to the experiments in Section \ref{subsec_transfer}, to assess the transferability of adversarial examples, we first use the targeted caption method with log-prob loss to find adversarial examples for 1,000 images in Show-and-Tell model (model $A$) with different $c$. 
We then transfer successful adversarial examples, i.e., the examples that generate the exact target captions on model $A$, to Show-Attend-and-Tell model (model $B$). The generated captions by model $B$ are recorded for transferability analysis. The results for transferability using log-prob loss is summarized in Table~\ref{tab:nologits_transfer}. The definitions of \textbf{tgt}, \textbf{ori} and \textbf{mis} are the same as those in Table~\ref{tab:transfer}. Comparing with Table~\ref{tab:transfer} ($C=1000, \epsilon=10$), the log probability loss shows inferior \textbf{ori} and \textbf{tgt} values, indicating that the additional parameter $\epsilon$ in the logits loss helps improve transferability.

\section{Attention on Original and Transferred Adversarial Images}
\begin{figure}[htbp]
		\centering
		\begin{subfigure}[b]{\linewidth}
			\includegraphics[width=1.0\textwidth]{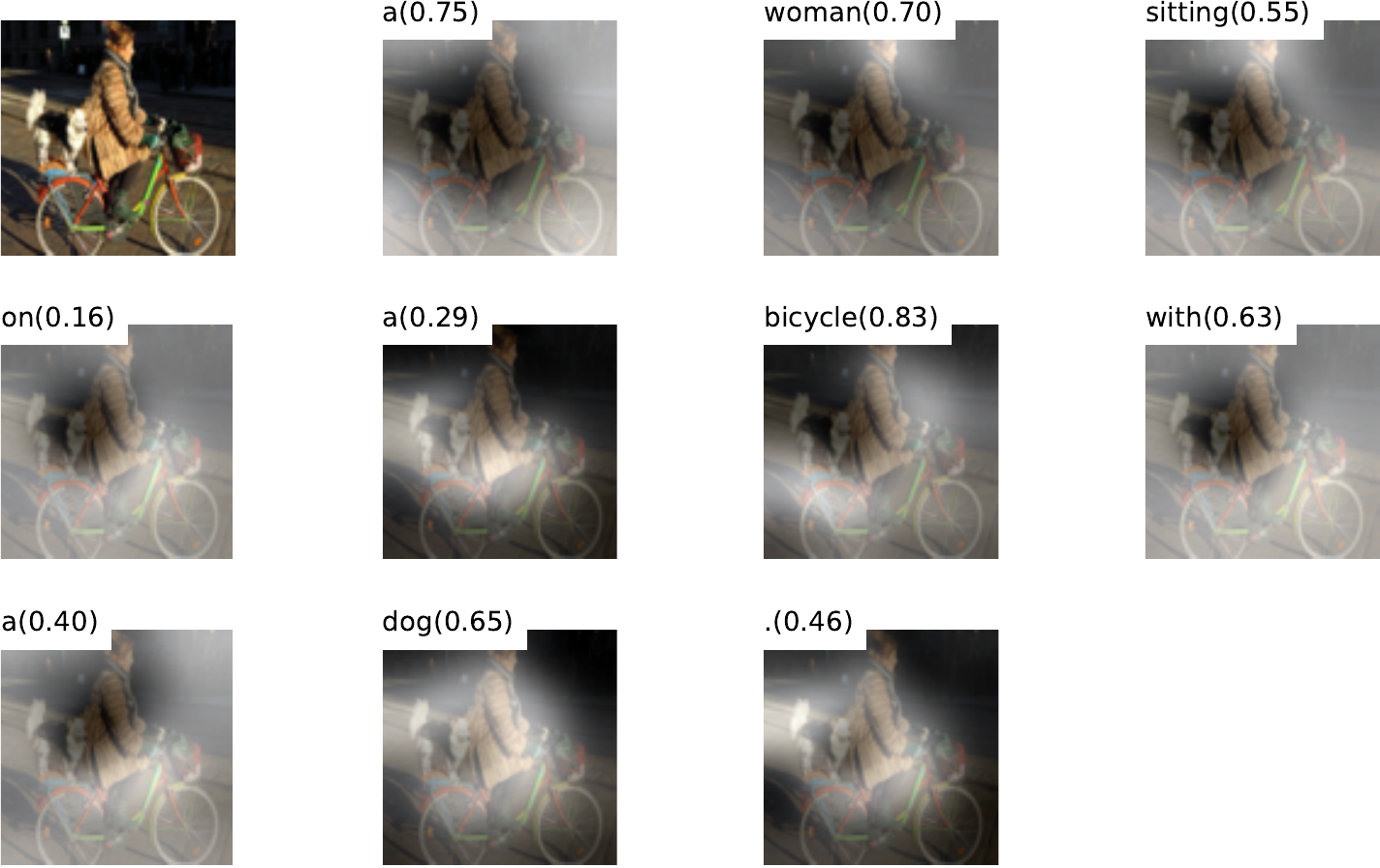}
			\caption{Original Image of Figure \ref{Fig_womanbike}}
		\end{subfigure}
        \\
		\centering
		\begin{subfigure}[b]{\linewidth}
			\includegraphics[width=1.0\textwidth]{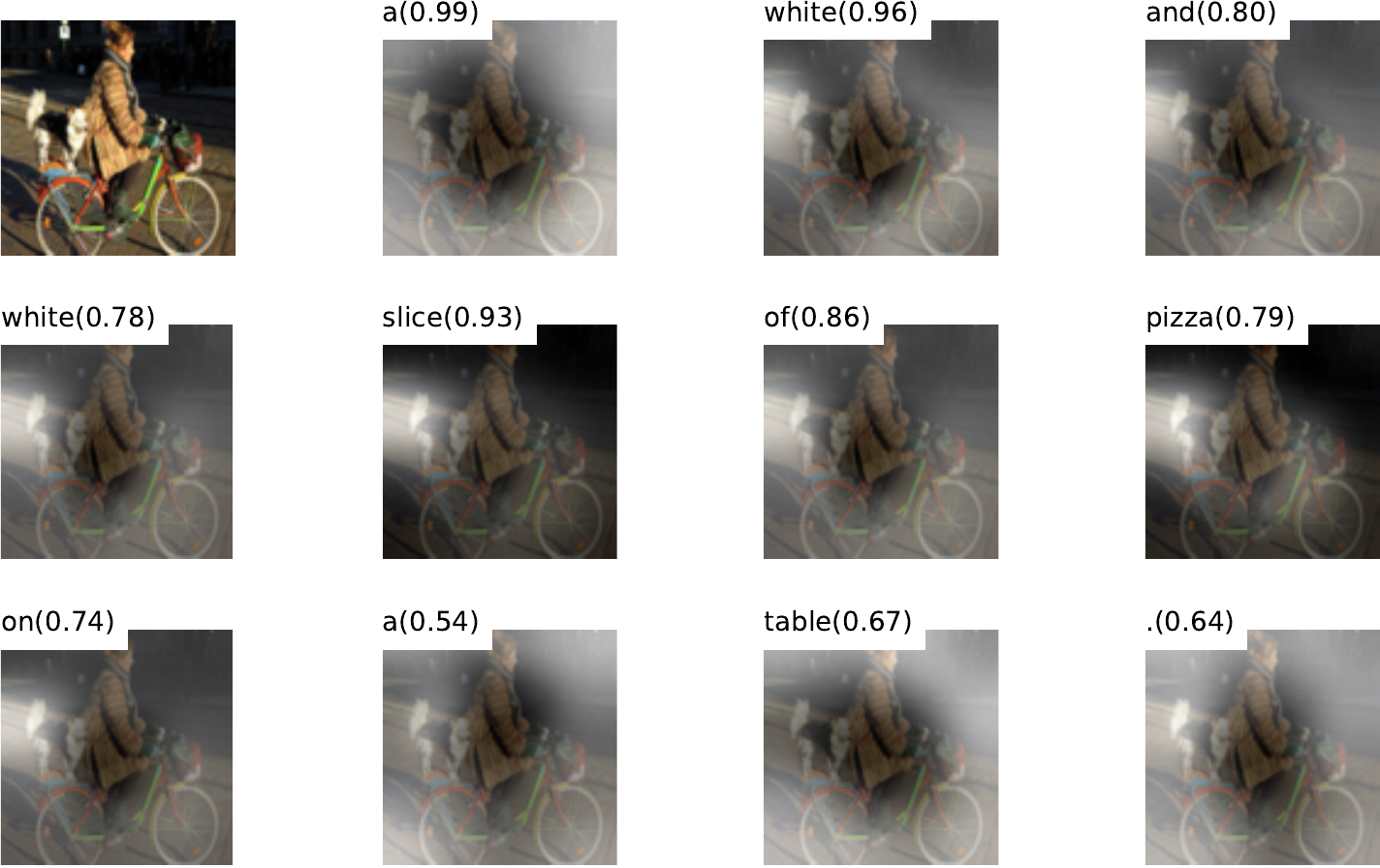}
			\caption{Adversarial Image of Figure \ref{Fig_womanbike}}
		\end{subfigure}
		\caption{Original and transferred adversarial image's attention over time on Figure \ref{Fig_womanbike}. The highlighted area shows the attention change as the model generates each word.}
        \label{Fig_bike_att}
	\end{figure}

\begin{figure}[t]
		\centering
		\begin{subfigure}[t]{\linewidth}
			\includegraphics[width=1.0\textwidth]{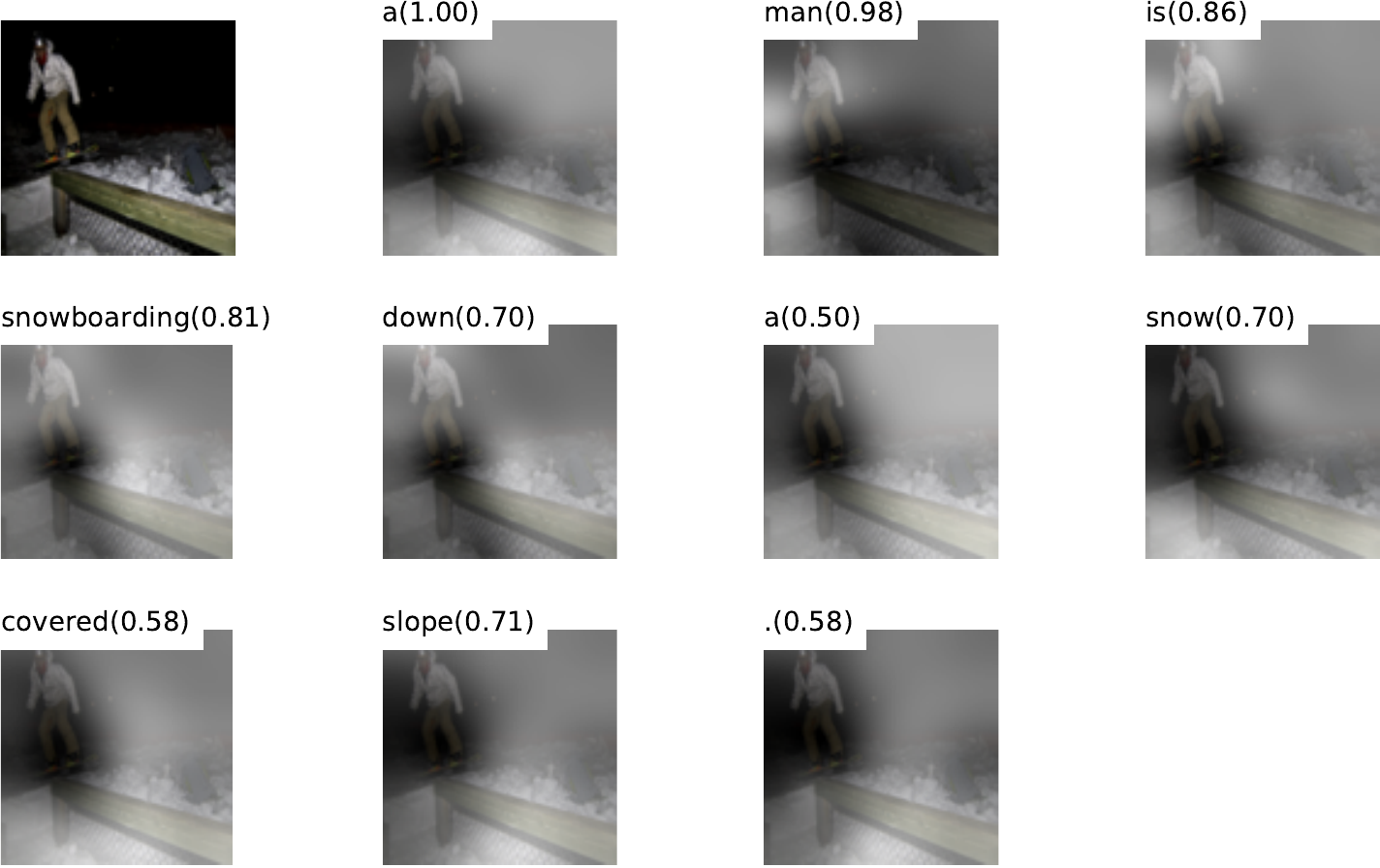}
			\caption{Original Image of Figure \ref{Fig_snowboard}}
		\end{subfigure}
        \\
		\centering
		\begin{subfigure}[t]{\linewidth}
			\includegraphics[width=1.0\textwidth]{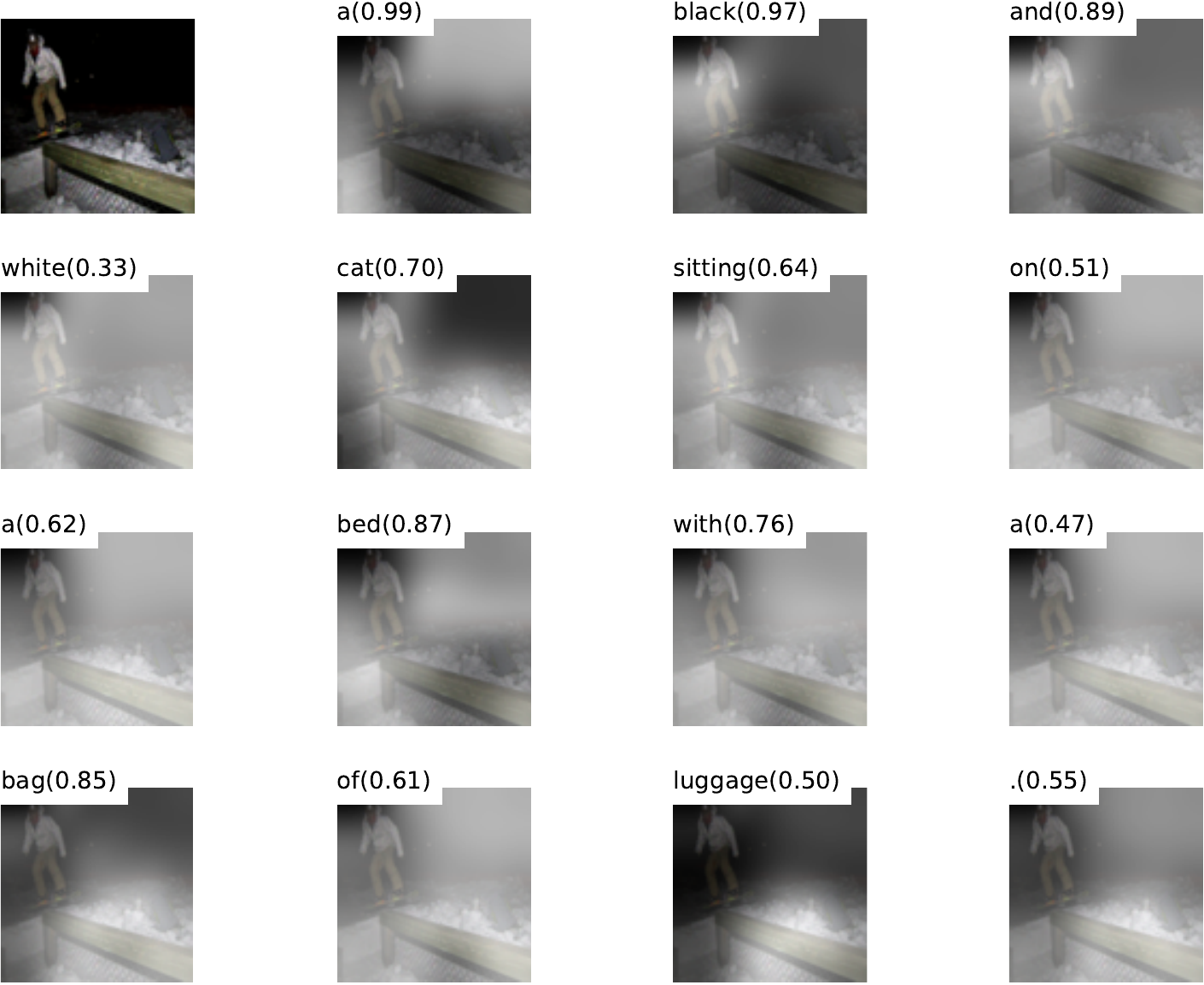}
			\caption{Adversarial Image of Figure \ref{Fig_snowboard}}
		\end{subfigure}
		\caption{Original and transferred adversarial image's attention over time on Figure \ref{Fig_snowboard}. The highlighted area shows the attention change as the model generates each word.}   
        \label{Fig_snowboard_att}
	\end{figure}

\begin{figure}[t]
		\centering
		\begin{subfigure}[t]{\linewidth}
			\includegraphics[width=1.0\textwidth]{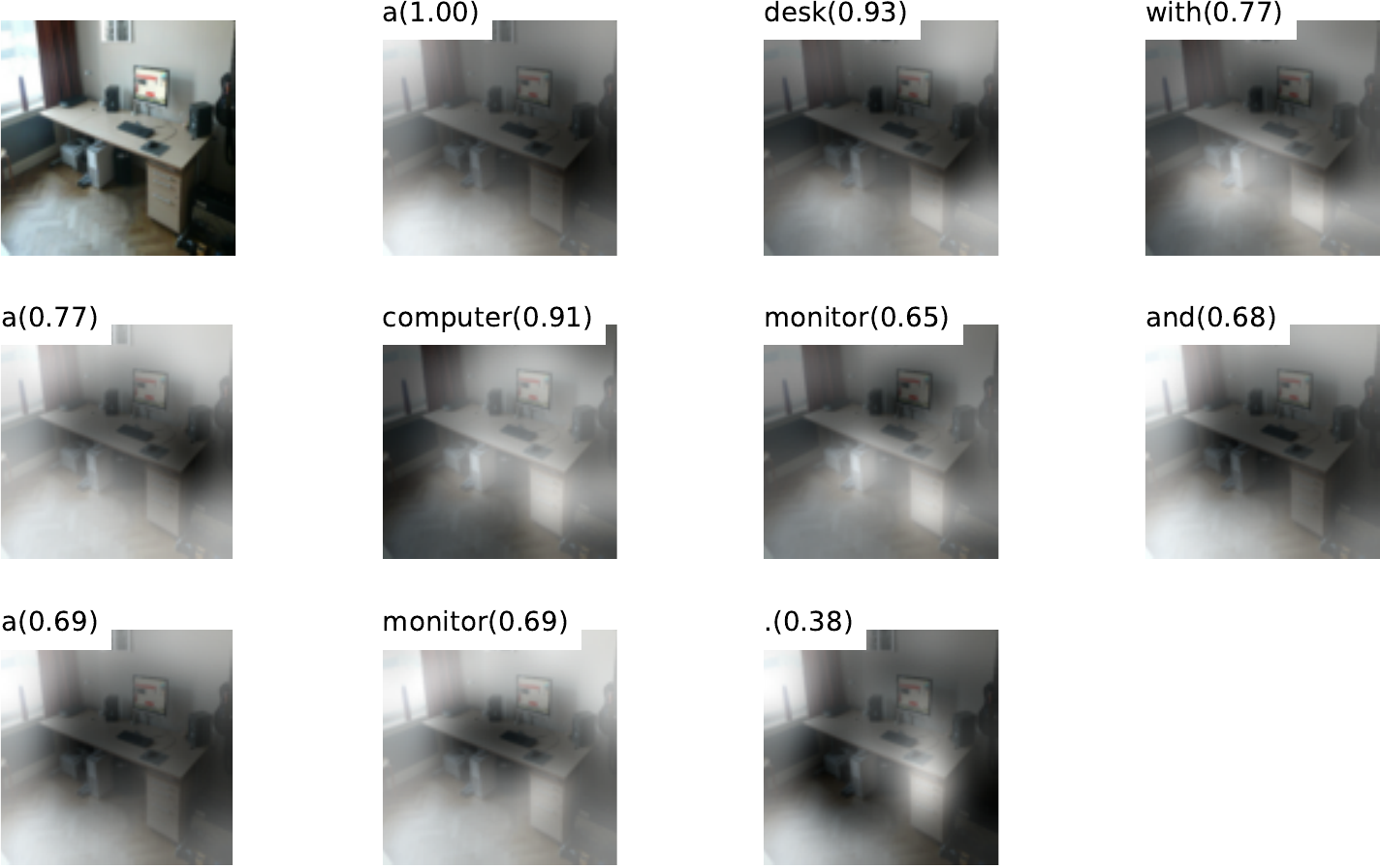}
			\caption{Original Image of Figure \ref{Fig_desk}}
		\end{subfigure}
        \\
		\centering
		\begin{subfigure}[t]{\linewidth}
			\includegraphics[width=1.0\textwidth]{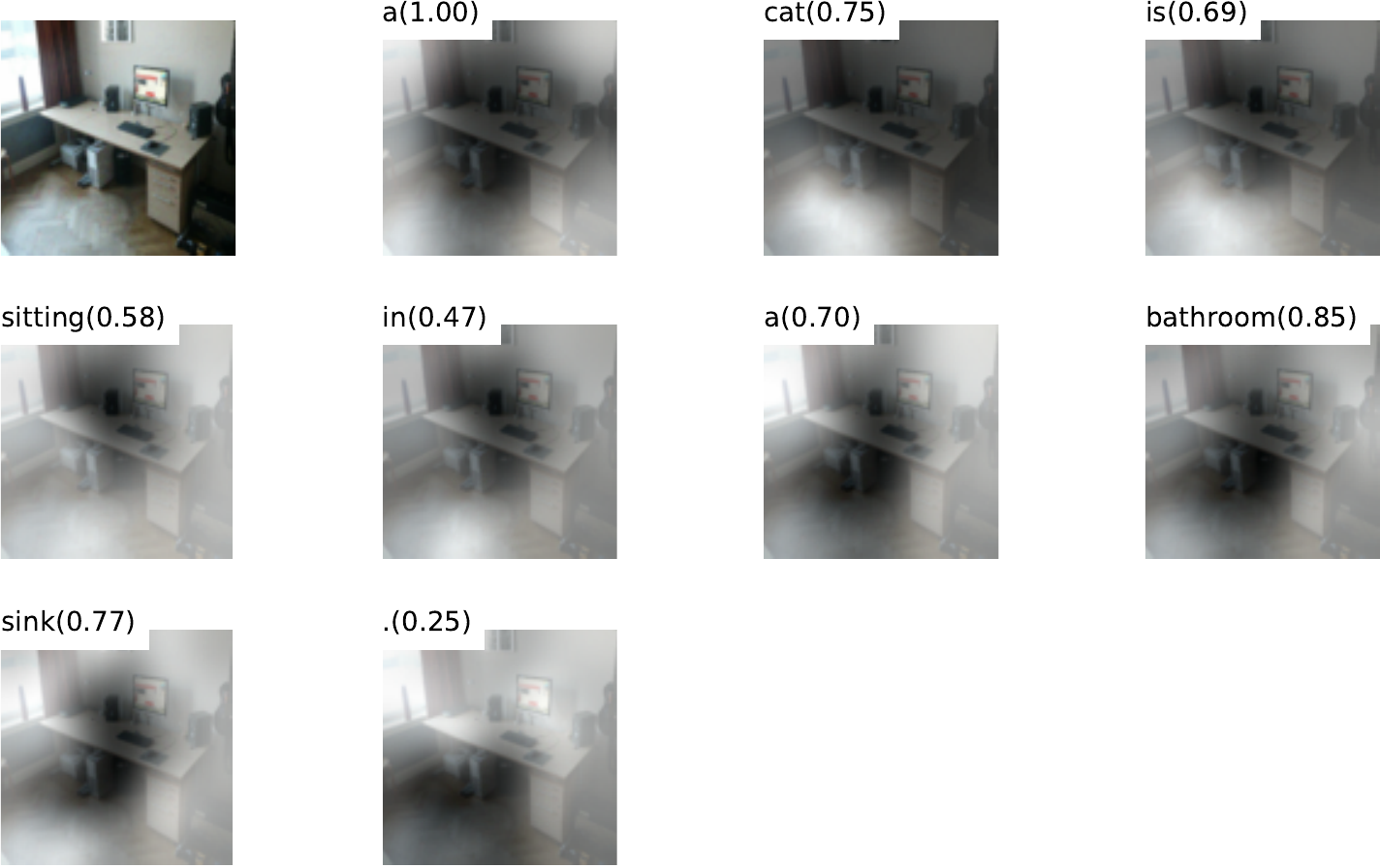}
			\caption{Adversarial Image of Figure \ref{Fig_desk}}
		\end{subfigure}
        \vspace{1.43cm}
		\caption{Original and transferred adversarial image's attention over time on Figure \ref{Fig_desk}. The highlighted area shows the attention change as the model generates each word.}   
        \label{Fig_desk_att}
	\end{figure}

Figures~\ref{Fig_bike_att}, \ref{Fig_snowboard_att} and \ref{Fig_desk_att} show the original and adversarial images' attentions over time. In the original images, the Show-Attend-and-Tell model's attentions align well with human perception. However, the transferred adversarial images obtained on Show-and-Tell model yield significantly misaligned attentions.

\newpage
\input{nologits_transferability}

\newpage

\newpage

%% file: nologits_transferability.tex
\begin{table}[t]
\vspace{-13.1mm}
\centering
\caption{Transferability of adversarial examples from Show-and-Tell to Show-Attend-and-Tell, using different $c$. Unlike Table~\ref{tab:transfer}, the adversarial examples in this table are found using the \textit{log-prob} loss and there is no parameter $\epsilon$. Similarly, a smaller \textbf{ori} or a larger \textbf{tgt} value indicates better transferability.}
\label{tab:nologits_transfer}
\scalebox{0.8}{
\begin{tabular}{cll|ll|ll|l}
\cline{2-8}
\bf
                                          &\multicolumn{2}{c|}{C=10} & \multicolumn{2}{c|}{C=100} & \multicolumn{2}{c|}{C=1000}&         \\ \cline{2-8} 
                                          & \bf{ori}  & \bf{tgt}     & \bf{ori}    & \bf{tgt}     & \bf{ori}    & \bf{tgt}     & \bf{mis}\\ \hline
\multicolumn{1}{l|}{\!\!BLEU-1}           &  .540     &  .391        &  .442       &  .435        & \bf .374    & \bf .500     & .657    \\        
\multicolumn{1}{l|}{\!\!BLEU-2}           &  .415     &  .224        &  .297       &  .280        & \bf .217    & \bf .357     & .529    \\        
\multicolumn{1}{l|}{\!\!BLEU-3}           &  .335     &  .143        &  .218       &  .193        & \bf .137    & \bf .268     & .430    \\        
\multicolumn{1}{l|}{\!\!BLEU-4}           &  .280     &  .101        &  .170       &  .142        & \bf .095    & \bf .207     & .357    \\ \hline 
\multicolumn{1}{l|}{\!\!ROUGE}            &  .525     &  .364        &  .430       &  .411        & \bf .362    & \bf .474     & .609    \\ \hline 
\multicolumn{1}{l|}{\!\!METEOR}           &  .240     &  .132        &  .179       &  .162        & \bf .135    & \bf .209     & .303    \\ \hline 
\multicolumn{1}{l|}{\!\!$\|\delta\|_2$}   &\multicolumn{2}{c|}{2.433}& \multicolumn{2}{c|}{4.612} & \multicolumn{2}{c|}{10.88} &         \\ \hline
\end{tabular}
}
\end{table}